\pdfoutput=1
\documentclass[12pt,a4paper]{article}
\usepackage[T1]{fontenc}
\usepackage{lmodern}
\usepackage[english]{babel}
\usepackage{graphicx}
\usepackage{caption}
\usepackage{amsmath}
\usepackage[toc]{appendix}
\usepackage{amsfonts}
\usepackage[square, numbers]{natbib}
\usepackage[x11names]{xcolor}
\usepackage{booktabs}
\usepackage{colortbl}
\usepackage{array}
\usepackage{makecell}
\usepackage{authblk}
\usepackage{listings}
\usepackage{wrapfig}
\usepackage{rotating}
\usepackage{natbib}
\usepackage{adjustbox}
\usepackage[colorlinks, citecolor=blue]{hyperref}

\title{The Roles of Contextual Semantic Relevance Metrics in Human Visual Processing}%\fsootnote{This manuscript has been submitted to \textit{Nature 
\date{\vspace{10ex}}

\author[1]{\footnotesize Kun Sun\thanks{\texttt{sharpksun@hotmail.com}}}
\author[2]{\footnotesize Rong Wang}
\affil[1]{\footnotesize Tübingen, Germany}
\affil[2]{\footnotesize Institute of Natural Language Processing, Stuttgart University, Stuttgart, Germany}

\begin{document}

\maketitle

\begin{abstract}

Semantic relevance metrics can capture both the inherent semantics of individual objects and their relationships to other elements within a visual scene. Numerous previous research has demonstrated that these metrics can influence human visual processing. However, these studies often did not fully account for contextual information or employ the recent deep learning models for more accurate computation. This study investigates human visual perception and processing by introducing the metrics of contextual semantic relevance. We evaluate semantic relationships between target objects and their surroundings from both vision-based and language-based perspectives. Testing a large eye-movement dataset from visual comprehension, we employ state-of-the-art deep learning techniques to compute these metrics and analyze their impacts on fixation measures on human visual processing through advanced statistical models. These metrics could also simulate top-down and bottom-up processing in visual perception. This study further integrates vision-based and language-based metrics into a novel combined metric, addressing a critical gap in previous research that often treated visual and semantic similarities separately. Results indicate that all metrics could precisely predict fixation measures in visual perception and processing, but with distinct roles in prediction. The combined metric outperforms other metrics, supporting theories that emphasize the interaction between semantic and visual information in shaping visual perception/processing. This finding aligns with growing recognition of the importance of multi-modal information processing in human cognition. %The study demonstrates that contextual semantic relevance metrics can predict both human visual processing and language comprehension/production, suggesting a unified approach to predicting human multi-modal information processing. 
These insights enhance our understanding of cognitive mechanisms underlying visual processing and have implications for developing more accurate computational models in fields such as cognitive science and human-computer interaction.

\end{abstract}

\textbf{Keywords:}	fixations, semantic metrics, vision language models, contextual information, multi-modality; top-down processing

\newpage 

\section{Introduction}
 
Human visual perception and processing are complex phenomena influenced by a variety of factors, both intrinsic and extrinsic. %The human visual system is designed to interpret and make sense of the vast amount of visual information we encounter daily. %This process involves several stages, including the detection, organization, and interpretation of visual stimuli.
One of the primary factors influencing visual perception is the human visual system as an intrinsic factor. The human visual system is designed to interpret and make sense of the vast amount of visual information we encounter daily. Visual perception is also affected by cognitive processes, and these processes are guided by past experiences, expectations, and the context in which visual stimuli are encountered \citep{enge2023instant}. For example, Gestalt principles suggest that the human brain tends to group similar elements together to form a cohesive picture of the world \citep{kovacs1993closed}.

Extrinsic factors in visual processing are mainly involved in image features. For instance, high contrast and distinct colors aid interpretation. Conversely, visual complexity, low contrast, and ambiguous elements hinder processing by slowing recognition and increasing cognitive effort. These factors collectively influence the speed and accuracy of visual interpretation. These basic extrinsic factors can be summarized as proportion of an object, saliency of an object, and the semantic relation between object and its context.

First, the proportion of an object within an image significantly influences human visual processing by affecting how the object is perceived and recognized. Larger objects or those occupying a significant portion of the visual field tend to be processed more efficiently. This is because larger objects are more likely to engage more extensive neural resources, facilitating quicker identification and interpretation (\citealp{dicarlo2012does}; \citealp{meese2023object}). Second, the saliency of an object in an image plays a crucial role in human visual processing by directing attention and influencing perception. Saliency refers to the distinctiveness of an object within its environment, often determined by low-level visual properties such as color, intensity, and orientation contrast. These salient features are processed rapidly, allowing the most prominent objects to be detected first %This mechanism is advantageous as it enables quick identification of behaviorally relevant stimuli, such as potential threats or important objects, by capturing attention almost immediately upon viewing a scene
 (\citealp{itti2001computational}; \citealp{parkhurst2002modeling}; \citealp{bruce2009saliency}; \citealp{fine2009visual}; \citealp{kamkar2018early}). 

%Although the proportion and saliency could influence human visual processing, human attention is strongly influenced by cognitive knowledge structures that represent the viewer’s understanding of the scene and of the world. Attentional priority is influenced by stored semantic representations of the relationships between the scene category and the objects it contains, along with the goals of the viewer. 

Cognitive knowledge structures, rather than visual features alone, play a dominant role in guiding human attention in visual processing (\citealp{antes1974time}; \citealp{loftus1978cognitive}; \citealp{wu2014guidance}). %Our mental models of the world and specific scenes significantly influence where we focus. 
Visual aspects such as proportion and saliency do impact perception. However, our stored semantic associations representing the understanding of the scene and of the world play a crucial role in visual processing. These associations link scene categories with expected objects, combined with our current objectives, largely determine attentional priorities in visual processing (\citealp{henderson2003human}; \citealp{hayhoe2005eye}). In essence, our brain's interpretation of a scene, based on prior knowledge and goals, frequently supersedes raw visual input in directing our attention. For instance, semantic information significantly influences visual processing, providing context and meaning to visual inputs. It facilitates quicker processing and changes perception of unfamiliar objects (\citealp{enge2023instant}). Semantic knowledge could guide visual attention and memory, improving working memory performance for familiar objects \citep{starr2020semantic} and directing attention to relevant scene elements \citep{wu2014guidance}. This interaction between semantic and perceptual information enhances our understanding of visual environments (\citealp{victoria2019relative}).  Semantic information can instantly alter visual perception, as demonstrated by changes in ERPs when unfamiliar objects are presented with functional descriptions. This indicates that semantic knowledge enhances real-time detection and interpretation of visual information by providing meaningful context (\citealp{wu2014guidance}; \citealp{enge2023instant}).

Due to the importance of semantic information on objects, particularly the semantic relationship between an object and its context in a scene, recently some computational metrics regarding such semantic relations have been proposed to predict and interpret human visual processing (mostly attention in real-world scenes). Such semantic metrics could fully represent semantic information on one object and the semantic relationship between the object and the scene objects. For example, \citet{hwang2011semantic} and \citet{hayes2021looking} used language-based semantic similarity (i.e., word of the object vs. the words of the surrounding objects) between the target object and the surrounding objects as a metric to demonstrate that such a semantic relationship influences attention in visual processing. On the other hand, without using language, researchers evaluated image features like color, shape, and size to estimate the visual similarity between an object and other objects as such semantic metrics (\citealp{rosch1976basic}; \citealp{duncan1989visual}). In other words, researchers employed visual information to estimate semantic metrics to study this phenomenon \citep{brust2018not}. Such semantic metrics based on visual information also show that they have great impact on human visual processing. 

Semantic and visual similarity are two distinct approaches designed to estimate the semantic relationship between an object and the scene objects (or the scene), despite sharing the common objective of quantifying this relationship. Whether leveraging linguistic information or visual cues, these metrics have proven invaluable in predicting and interpreting human visual perception and processing. Early research relied on human ratings to estimate these metrics. However, as the amount of stimuli increased and the need for precision grew, computational methods gradually replaced human-based assessments. These metrics now serve as a foundation for developing computational models that emulate human visual cognition. Although these metrics are quite promising to explore human visual mechanisms, there is great room for improvement. For instance, these metrics ever proposed seemingly have not completely incorporated the contextual information. The advent of advanced deep learning techniques in vision and language processing, particularly large language models and multi-modal models, offers opportunities to compute these semantic metrics more conveniently and precisely. Currently, language-based and vision-based metrics are often investigated separately. This raises questions about the distinct roles semantic and visual metrics play in human visual processing, and the potential benefits of merging these two types of metrics. Further, we could employ more sophisticated statistical methods to explore how these metrics influence human attention in visual processing. In other words, this study proposes new computational methods to calculate more powerful and interpretable semantic metrics for predicting and explaining human visual processing. These advanced metrics could provide deeper insights into the complex interplay between visual input, semantic understanding, and attentional processes in human cognition. %We could distinguish two methods as semantic similarity and visual similarity although both have the same goal of computing the semantic relationship betwen object and other objects. Regardless using language or vision  information to compute semantic relationship between an object and other objects, such metrics on visual similarity could predict and interpret human visual perception and processing and further could help build up interpretable computational models for human visual processing. Although these metrics are quite promising to explore human visual mechanisms, there is great room for improvement. For instance, these metrics ever proposed seemingly have not completely incorporated the contextual information. The current study proposes a new method to compute more powerful and interpretable semantic and visual similarity which can incorporate the contextual information more comprehensively.   

%Moreover, language could shape visual perception by influencing categorization and interpretation of stimuli. Language affects both higher-level recognition and lower-level discrimination processes, leading to more categorical perception (\citealp{noorman2018words}; \citealp{lupyan2020effects}).

Eye movement have been extensively employed to investigate human visual processing. The measures like total fixation duration and fixation number tend to indicate object processing difficulty. Longer durations and more fixations suggest higher cognitive effort (\citealp{ramamurthy2015human}; \citealp{negi2020fixation}). Complex or unfamiliar stimuli often lead to longer fixations, while multiple fixations may indicate object complexity or ambiguity \citep{cronin2020eye}. These patterns provide insights into visual perception and attention allocation (\citealp{cronin2020eye}; \citealp{negi2020fixation}), offering a window into the visual system's processing challenges. We are interested in using eye-tracking databases as testing datasets to evaluate the effectiveness and validity of the metrics proposed in the present study. Additionally, the datasets on human visual perception and processing also face several weaknesses and potential problems, which can impact the comprehensiveness and applicability of findings in this field. First, some related research was merely involved in simplistic task. Humans actually process much more complicated visual scenario in real life. Second, effective real-world visual processing requires a comprehensive understanding of the entire scene. To assess this ability, researchers should ask participants to verbalize captions that capture the overall theme of an image. This approach helps evaluate whether participants have truly grasped the essence and context of the visual stimulus. Addressing these weaknesses requires some datasets from naturalistic visual processing with visual understanding. %Third, the metrics of semantic or visual similarity could be optimized to incorporate more contexutal information from the vision due to the the rapid development of deep learning models. The relationship between (language-based) semantic similarity and (vision-based) visual similarity has not been explored. Is it possible to integrate two types of metrics?  

To address the limitations of previous research, this study leverages recent deep learning techniques to compute a number of new metrics on semantic relationship between an object and others. We used a large eye-movement dataset from naturalistic visual processing. To better understand the impacts of these metrics on eye-movement measures in the eye-movement dataset, we  applied a generalized additive mixed-effects model to statistically explore the relationships between eye-movement measures and the proposed metrics. Through advanced statistical analysis, we aim to investigate the roles of these metrics in human visual processing, ultimately enhancing our understanding of the cognitive mechanisms underlying visual processing. Moreover, this study not only addresses the limitations of previous research but also contributes to a broader understanding of the intricate interplay between visual and language processing.

\section{Related Work}

\subsection{Factors influencing visual perception and processing}
Visual perception is influenced by a variety of factors that affect how objects are recognized and processed by the human brain. These factors encompass both cognitive and environmental elements, contributing to the complexity of visual processing.

\textbf{Contextual Information}: Contextual information plays a pivotal role in object recognition, as highlighted by the recognition-by-components theory. It emphasizes the interaction between structural components and their configuration within the context of the whole, suggesting that context is crucial in how objects are perceived and recognized (\citealp{albright2002context}; \citealp{Lee2009context}; \citealp{Lauer2021context}). For instance, objects are often recognized more accurately when presented in a semantically consistent scene, as the surrounding context provides additional cues that facilitate recognition. Studies have shown that even without explicit spatial scene structure, contextual materials can affect object processing, indicating the importance of both spatial and material context in visual perception \citep{carmichael1932experimental}.

%\textbf{Object Familiarity}: Familiarity with an object significantly influences visual perception. Research indicates that prior knowledge about an object can affect the brain pathways used for processing visual information, thereby influencing recognition and processing efficiency. Familiar objects are processed more efficiently because the brain can leverage existing memory representations, leading to faster and more accurate recognition. This familiarity effect underscores the role of experience and memory in shaping visual perception, as familiar objects are often recognized with greater ease and less cognitive effort.

\textbf{Visual Salience}: Visual salience refers to the prominence of certain features, such as contours, colors, and textures, that make objects stand out in a visual scene. These features contribute to the visual salience of objects, affecting how they are detected and processed. Salient features capture attention more readily, guiding the viewer's focus and facilitating object recognition. The significance of visual salience is well-documented, as it plays a critical role in directing attention and enhancing the perceptual processing of objects in complex visual environments.

In the recent years, models based on image salience have provided the most influential approach to visual attention (\citealp{itti2001computational}; \citealp{parkhurst2002modeling}). These classic saliency models propose that attention is controlled by contrasts in primitive, presemantic image features such as luminance, color, and edge orientation
(\citealp{treisman1980feature}; \citealp{wolfe2017five}). Although theories based on image salience can account for key data regarding attentional guidance, it is also clear that in meaningful real-world scenes, human attention is strongly influenced by cognitive
knowledge structures that represent the viewer’s understanding of the scene and of the world (\citealp{antes1974time}; \citealp{loftus1978cognitive}; \citealp{henderson2018meaning}).

%\textbf{Object Complexity}: The complexity of an object's shape and features is another factor influencing recognition. Studies suggest that recognition involves multiple systems that may represent objects by combinations of views or structural primitives. Complex objects may require more cognitive resources for processing, as the brain must integrate various features and viewpoints to form a coherent representation. This complexity influences processing by necessitating more intricate neural computations, highlighting the interplay between object features and perceptual mechanisms in visual recognition.

 \textbf{Position and Proportion of Objects}: The position and proportion of objects in an image significantly affect object processing in humans \citep{Ayzenberg2024visual}. Research indicates that object recognition is adapted to positional regularities found in the natural environment. Objects typically appear in expected locations relative to the visual space and other objects, such as an airplane being seen in the upper part of a scene. These positional regularities help facilitate object detection and recognition. This adaptation to natural positional patterns enhances the efficiency and accuracy of visual perception, reflecting the brain's ability to learn and exploit environmental regularities for improved cognitive processing. 
 
 Conversely, smaller objects or those occupying a minor portion of the image may pose challenges for visual processing. They can be more difficult to detect and identify, especially if they are surrounded by other visual stimuli or if the overall scene is complex. The visual system may require additional cognitive effort to focus on these smaller objects, potentially leading to slower recognition times. This is particularly relevant in scenarios where multiple objects are present, and attention must be selectively directed to the most relevant features. In this way, the proportion of an object in an image plays a critical role in determining the efficiency and accuracy of visual processing (\citealp{dicarlo2012does}; \citealp{meese2023object}). 
 
 %In short, due to the fundamental roles of salience and proposition demonstrated by numerous studies,  the present study took them as control predictors in the statistical analysis.  The position of objects could be taken as a random variable because of its categorical fator.  
 Due to the established roles of salience and proportion in human visual processing, the current study incorporates them as control predictors in our statistical analysis. We treat object position as a random variable because of its categorical nature, allowing us to account for variability while focusing on our primary variables of interest. This approach helps isolate the effects of the semantic and visual similarity metrics under investigation.

\subsection{Semantic or visual relevance}

  In the section of introduction, we have introduced the role of semantic information in human visual processing.  Here we want to detail how semantic information afffect neural activity in visual perception. For instance, semantic information can affect the neural activity of visual processing as early as 100-150ms after stimulus onset, as evidenced by changes in the P1 component of event-related potentials (ERPs). The N170 ERP component (150-200ms) shows larger amplitudes for semantically informed perception, suggesting that semantic information influences higher-level visual perception and object recognition processes. Later stages of processing (400-700ms) are also affected, as shown by reduced N400 amplitudes and changes in alpha/beta band power for semantically informed perception \citep{shoham2024using}. Importantly, these effects can occur instantly after semantic information is provided, even for previously unfamiliar objects, without requiring extensive learning or experience. This suggests that semantic knowledge can rapidly and dynamically influence visual perception in a top-down manner (\citealp{de2016meaning}; \citealp{enge2023instant}).
  
%Due to the importance of semantic information on objects in human visual processing, some computational metrics regarding such semantic relations have been proposed to predict and interpret human visual processing (mostly attention in real-world scenes). Such semantic metrics could fully represent semantic information on one object and the semantic relationship between the object and the scene objects. These semantic metrics have been tested to show that they play a cruicial role in human visual processing. There are two appraoches to estimating such metrics: language and vision.  
Semantic information plays a crucial role in human visual processing, prompting researchers to develop computational metrics that quantify semantic relationships between objects and their contexts in real-world scenes. These metrics aim to represent both individual object semantics and their relationships to other scene elements, proving pivotal in predicting and explaining human visual attention mechanisms. Two primary approaches have emerged for estimating these semantic metrics: language-based methods, and vision-based methods. Both approaches offer unique insights into how semantic information shapes our visual experience, contributing significantly to our understanding of human visual cognition and attention allocation in complex, naturalistic environments.

 Research on semantic similarity's effect on human visual processing reveals a complex and significant relationship. Studies have shown that semantically related objects in a scene tend to capture more attention, with regions containing objects semantically similar to the overall scene category or to other objects receiving greater focus \citep{jiang2022visual}. However, this relationship is nuanced; while semantic similarity can guide attention, it may also require more cognitive effort to process, especially as the number of objects increases\citep{garcia2020visual}.

 Similarly, visual similarity plays a fundamental role in human visual perception and processing, influencing various aspects of visual cognition from low-level perception to high-level object and scene recognition. Research has shown that visually similar elements tend to be grouped together in perceptual organization, guiding attention allocation and facilitating object and scene recognition by allowing efficient comparison with stored mental representations. This similarity-based processing affects memory encoding and retrieval, with visually similar items often encoded and recalled together. In visual search tasks, the degree of similarity between targets and distractors significantly impacts search efficiency, while experience with similar stimuli can enhance perceptual learning and fine-grained discrimination abilities. Visual similarity also contributes to categorical perception, aesthetic judgments, and elicits similar patterns of neural activity for related stimuli. Importantly, visual similarity often interacts with semantic similarity, jointly shaping our perception and understanding of visual scenes. This complex interplay between visual similarity and various cognitive processes underscores its crucial role in human visual cognition, providing insights that are valuable for developing more accurate models of human vision and improving computer vision algorithms to better align with human visual processing.
 
 Further, some research explored the relation between semantic and visual similarity. For instance, there is a moderate correlation between visual and semantic similarity, with semantic information providing insights beyond basic categorical distinctions. This interplay influences various cognitive processes, including visual search, object recognition, and memory encoding \citep{zelinsky2013modeling}. The effect is particularly pronounced when focusing on object categories rather than backgrounds, suggesting a close tie to object recognition processes \citep{wu2014guidance}. These findings highlight the crucial role of semantic information in shaping human visual perception and attention, demonstrating that our understanding and processing of visual scenes are deeply influenced by the semantic relationships between objects and their context. Moreover, \citet{deselaers2011visual} explore the correlation between visual and semantic similarity through the ImageNet dataset. The relationship is complex and multifaceted. Different visual and semantic similarity measures show varying degrees of correlation, suggesting that the connection is not straightforward. For example, semantic similarity provides more information about visual similarity than basic categorical distinctions \citep{rosch1976basic}. WordNet semantic similarity carries more information about visual similarity \citep{brust2018not}. The relationship varies depending on the specific concepts and context. For instance, some concepts may be semantically similar but visually different (e.g. an orchid and a sunflower), while others may be visually similar but semantically distant (e.g. a deer and a forest) \citep{brust2018not}. Further, visual and semantic similarities interact in complex ways in human perception. \citet{jiang2023effect} found that regions containing objects semantically related to the overall scene category or to other objects tend to capture more attention.

 The primary challenge in assessing semantic similarity is that existing methods often fail to comprehensively incorporate context. A typical approach for predicting visual processing, as seen in \citet{hayes2021looking}, involves summing the similarity between the target object word and the words of surrounding objects. However, this method overlooks the relationship between the target object and the overall theme of the image. The collection of all objects in an image does not necessarily define its theme. The relationship between an object and the entire image (or other objects) is a crucial factor in image processing and object recognition, as various studies have highlighted. For example, in a kitchen scene, a ``knife'' contributes to the cooking theme but does not fully define it. Other elements like pots, ingredients, and a stove are necessary to establish the complete kitchen context. This example shows how individual objects support, but do not solely determine, an image's overall theme. The relationship between the knife and surrounding objects creates a comprehensive kitchen scene, highlighting the importance of considering multiple elements in image processing and object recognition.
 
 To address this, we can identify several metrics: the connection between the target object and the surrounding objects, the connection between the target object and the entire image, and a combined metric that includes both. These metrics can be computed using the image information and the corresponding language expressions. When vision-based and language-based metrics are combined, could this integrated metric more accurately predict human eye movements? Moreover, with the remarkable advancements in large language models, vision models, and multi-modal models, we are now better equipped to compute these metrics more effectively.

\subsection{Computing contextual semantic or visual relevance}

%Many studies have demonstrated the presence of the semantic plausibility benefit effect in reading and language comprehension. However, much of this research has relied on human rating methods, where participants assess semantic relatedness or relevance to determine plausibility (\citealp{hohenstein2010semantic}; \citealp{delong2014predictability}). While valuable, human ratings are inherently imprecise and can be costly, making them impractical for large-scale text analysis. As an alternative, semantic similarity measures, such as cosine similarity, offer a computational approach to predicting language processing by assessing how closely related the meanings of two words or phrases are. This method has been widely applied in fields like natural language processing (NLP), cognitive psychology, and artificial intelligence (see  \citealp{jabeen2020semantic}; \citealp{harispe2022semantic}). A substantial body of empirical research has shown that semantic similarity can effectively predict eye movements during reading and neural signals during language comprehension (\citealp{pereira2018toward}; \citealp{broderick2018electrophysiological}; \citealp{hale2022neurocomputational}). The primary advantage of semantic similarity lies in its ability to efficiently process large-scale texts, making it a more practical alternative to human rating methods.

Recent research has yielded effective methods for computing contextual semantic similarity (relevance) among words within a sentence. These approaches incorporate contextual information using an ``attention-aware'' technique, resulting in more robust contextual semantic relevance metrics (\citealp{sun2023interpretable}; \citealp{sun2024attention}). This new approach to calculating contextual semantic relevance for each word in a sentence has outperformed two existing methods: the \textit{cosine} method, modified from \citet{frank2017word}, and the \textit{dynamic} method from \citet{sun2023interpretable}, all trained on the same pre-trained word embedding database. The \textit{cosine} method proposed by \citet{frank2017word} only considered content words. A similar approach by \citet{michaelov2024strong}, which computes semantic relevance using the cosine between the vector of the target word and the mean vector across each word in the context, is essentially identical to \citet{frank2017word}. The limitations of these methods have been discussed in detail by \citet{sun2023interpretable} and \citet{sun2024attention}. To address these limitations, \citet{sun2023interpretable} introduced a modified \textit{cosine} approach that considers both content and function words in the context. For a given target word, this method concatenates the vectors of the three preceding words, regardless of their grammatical function. The cosine similarity is then calculated between the target word vector and the sum of the preceding three word vectors, yielding the \textit{cosine semantic similarity} value for the target word. While \citet{broderick2019semantic} proposed a Euclidean approach with variable context window sizes, this method is significantly affected by high-dimensional vectors and exhibits unstable performance. Although both the \textit{cosine} and \textit{Euclidean} approaches have been partially optimized, they still retain inherent weaknesses. Given these issues, researchers have opted to compare the metric computed by the modified \textit{cosine} method with newly proposed metrics, focusing on approaches that address the limitations of earlier methods and provide more reliable and context-sensitive measures of semantic relevance.

Clearly, the methods used to compute semantic relevance in human language processing cannot be directly applied to estimating the metrics for human visual processing. Despite this, we could tailor some exiting algorithms to compute the relevant metrics for visula processing. For instance, one could calculate the sentence semantic similarity between an object's name and the image's \texttt{caption}(i.e., one sentence or phrase) to estimate the semantic relevance between the object and the image. Alternatively, a metric could be derived by summing the semantic similarities between the object word and the words representing its surrounding objects. 

%Regarding visual similarity, we could employ some popular vision language models or multi-modal models (e.g., \texttt{CLIP} [Contrastive Language-Image Pre-training] by OpenAI, \texttt{Flamingo} by DeepMind, VisualBERT, etc.) to generate embeddings representing visual information on the target object and embeddings for the visual cue representing the scene. Using these embeddings can quantify the semantic relationship between a target object and the scene. Additionally, we can use these models to generate embeddings representing the visual information on the target object and its surrounding objects, and then calculate the cumulative semantic similarities between the object and its surrounding objects, providing a comprehensive measure of visual similarity. 

To quantify visual similarity, we propose employing state-of-the-art vision-language models or multi-modal models such as \texttt{CLIP}, \texttt{Flamingo}, and \texttt{VisualBERT}. These models can generate embeddings representing the visual information of both the target object and the scene. By comparing these embeddings, we can calculate the semantic relationship between an object and its context. Additionally, we can generate embeddings for the target object and its surrounding objects, allowing us to calculate cumulative semantic similarities. This approach provides a comprehensive measure of visual similarity, capturing both object-scene relationships and object-object interactions within the scene. By employing these advanced models, we can obtain a holistic measure of visual similarity that incorporates semantic understanding, going beyond simple feature matching to capture nuanced visual relationships in complex scenes.

We propose creating a new metric by linearly combining semantic and visual similarity scores. This approach effectively incorporates contextual information, enhancing the metric's relevance to visual processing. The methodology section provides a detailed explanation of how to compute these new contextual metrics, offering a more comprehensive way to analyze visual relationships in complex scenes.

\section{Materials and Methods}
\subsection{Testing datasets}
The ``Human Attention in Image Captioning'' dataset aims to provide a detailed understanding of how humans allocate attention when describing images \citep{he2019human}. This dataset includes 1,000 diverse images, each accompanied by raw data such as eye-fixations (e.g., total duration for an object in a image, and fixation number), verbal descriptions (the participants speak the ``caption'' for an image), and transcribed text descriptions from five native English speakers. The eye-fixation data captures where and for how long participants focus on different parts of an image, offering insights into the visual attention patterns that humans exhibit during the image captioning process. This information is crucial for understanding which parts of an image are deemed important by humans and how these areas influence the descriptive language used in captions.

The dataset also includes audio recordings of participants verbally describing the images, capturing the spontaneity and richness of natural language. These verbal descriptions are transcribed into text, facilitating analysis of the content and structure of the descriptions. By correlating eye-fixation patterns with elements of the verbal descriptions, researchers can study the relationship between visual attention and linguistic output. This dataset is valuable for various research applications, including improving image captioning models by integrating human attention patterns, developing models to predict human visual attention based on image content, and exploring the interaction between visual and linguistic modalities in human cognition.

\subsection{Computing vision-based semantic relevance}

The first semantic relevance metric is calculated using the cosine similarity between the embedding vectors of the whole image and the embedding of an individual object detected within that image. The \texttt{CLIP} model (\texttt{openai/clip-vit-large-patch14}) \citep{radford2021learning} is used to generate these embeddings, where the entire image's embedding, \(\mathbf{v}_{\text{image}}\), is compared to each detected object's embedding, \(\mathbf{v}_{\text{object}}\). The cosine similarity, \(\text{similarity} = \text{cosine\_similarity}(\mathbf{v}_{\text{image}}, \mathbf{v}_{\text{object}})\), quantifies the degree of similarity between the image and the object, with a value ranging from -1 to 1, where 1 indicates identical orientation, 0 indicates orthogonality, and -1 indicates opposite orientation. For instace, the object ``baby' in Fig.~\ref{fig:semrev} has its embedding generated by the \texttt{CLIP} model,  and the whole image also has its embedding, and after using the cosine method, we can get the cosine value, and the value is taken as the semantic relevance for the object ``baby''. The other objects such as ''bottle'' and ``lemmon'' is computed like this. % This similarity score is stored alongside other object-specific information such as fixation duration, object boundary, proportion of the image occupied, and position within the image. 
The similarity metric serves as an indicator of how well the object's visual features align with the overall image context, potentially aiding in tasks like object saliency detection or understanding visual attention patterns.

The other semantic relevance metric computes a score for a target object within an image by considering its resemblance to the entire image and its surrounding objects (``objs\_vissim'' = objects-based visual similarity). In other words, the metric is the extension of the first metric. For instance, we have a value for the semantic value between ``baby'' and the whole image. Meanwhile, we calcuated the other two cosine values between ``baby'' and any of the other two objects based on their embedding generated by the \texttt{CLIP} model. Ultimately, we added up all these cosine values to obtain a new value. As a matter of fact,  the method takes as input the target object's embedding vector, surrounding objects' embeddings with adjacency information, and the whole image's embedding vector. The function begins by calculating the cosine similarity between the target object's embedding and the whole image's embedding, \(\mathbf{v}_{\text{image}}\), denoted as \(\cos_0 = \text{cosine\_similarity}(\mathbf{v}_{\text{target}}, \mathbf{v}_{\text{image}})\). We continued to compute cosine similarity over each surrounding object, \(\cos_i = \text{cosine\_similarity}(\mathbf{v}_{\text{target}}, \mathbf{v}_{i})\), where \(\mathbf{v}_{i}\) is the embedding of the \(i\)-th surrounding object. %If the object is not adjacent, as determined by bounding box overlap, the similarity score is scaled down by a weight of 0.7. The weight plays  a role of reduced memory in humans because we believe that when the near objects are not adjacent with the target object, it is likely to a  representation of memory retention's decline over time, where information retention halves daily over several days. To mimic human memory decay, we assigned higher weights (the factor 1.0) to adjacent object and lower weights (e.g.,0.7, 0.5, 0.3 depending the distance and object number) to more distant ones. This algorithm parallels the human forgetting pattern with the attentional weights in our study.  
These similarity scores are accumulated as the final output.

In the broader context, object detection is performed using the \texttt{DETR} model (\texttt{facebook/detr-resnet-50}) \citep{carion2020eodt} to identify objects within an image, and the \texttt{CLIP} model is employed to generate embeddings for both the entire image and individual objects. The adjacency of objects is determined by checking for overlap in their bounding boxes. For each detected object, we computed a cumulative similarity score. This summed similarity serves as a metric for assessing how well a target object integrates within the context of the entire image and its surrounding objects more comprehensively, potentially aiding tasks such as object saliency detection, image captioning, or context-aware object recognition. The panel \textbf{B} of Fig.~\ref{fig:semrev} illustrates how these vision-based metrics are computed. 
%Moreover, these vision-based metrics effectively simulate the bottom-up processing hypothesis in visual perception. Bottom-up processing involves the analysis of sensory input based on inherent features such as color, shape, and contrast \cite{dijkstra2017distinct}. By incorporating these low-level visual features, our computational metrics capture the rapid, automatic processing of the bottom-up visual information. %This approach allows us to better understand how basic visual features contribute to attention allocation and object recognition in complex visual environments.

\subsection{Computing language-based semantic relevance}

% The following outlines how to compute these new attention-aware metrics.%, which take into account contextual information in a comprehensive and thorough manner.

The first method is based on image (vision)-generated embeddings. However, we can compute contextual semantic similarity based on language. This method is not related to graph-based embedding. Instead, it heavily relies on language and word embeddings.

\begin{figure*}[htp]%[!h]%[htp]
	\vskip -0.06in
	\centering
	
	\includegraphics[width=0.99\textwidth]{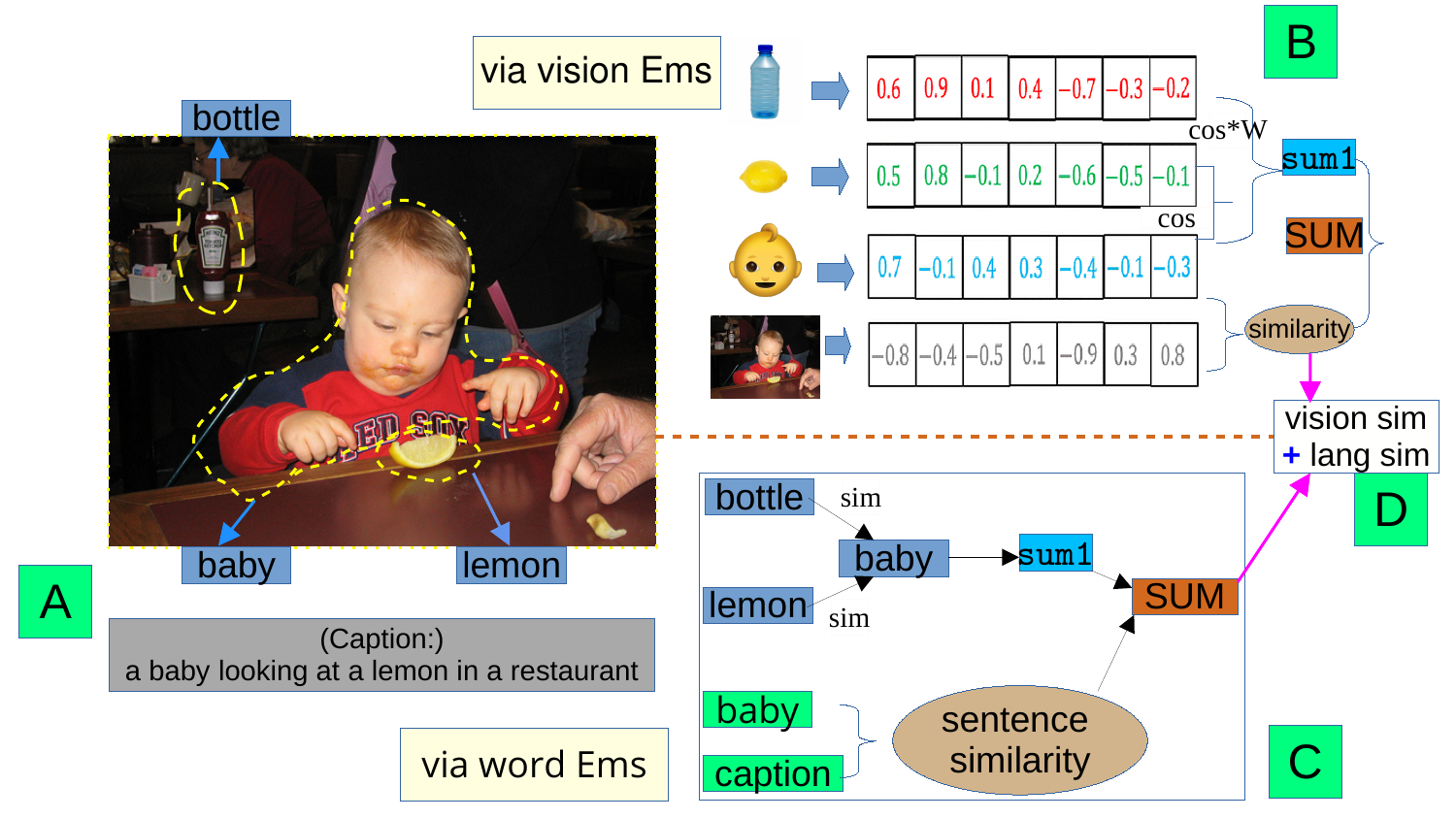}
	
	\caption{The computational method for contextual semantic relevance. Note: There are four panels. Panel A represents the image information, Panel B denotes the computation of vision-based metrics, Panel C illustrates the computation of language-based metrics, and Panel D shows the combination of one vision-based metric and one language-based metric. ``Ems''=embeddings}
	
	\label{fig:semrev}
	\vskip -0.11in	
\end{figure*}

The first metric, ``sent\_sim''(= sentence-based similarity), is calculated to measure how closely an object name (i.e., word) relates to the caption for an image. Typically, a caption for an image is a sentence or phrase. For instance, in Fig.~\ref{fig:semrev}, the target object has the name ``baby'' in language, and the caption for this image is ``a baby looking at a lemon in a restaurant''. At this point, we can treat both as two different sentences. By employing a sentence transformer model, specifically the \texttt{Sentence Transformer} from the \texttt{sentence-transformers} library \citep{reimers-2019-sentence-bert}, we encode the word ``baby'' into an embedding vector, \(\mathbf{v}_{\text{object}}\). The caption is then tokenized into individual sentences, each of which is encoded into its own embedding vector. The cosine similarity between the ``baby'' embedding and each sentence's embedding is computed using the cosine function, which measures the angle between the two vectors in the embedding space. The sentence similarity is defined as the maximum cosine similarity value obtained, indicating the sentence that is most semantically similar to the object name. This metric is useful for assessing the contextual relevance of object names within text, potentially aiding tasks such as image captioning or text-based object recognition.

\begin{wrapfigure}{l}{0.62\textwidth}
	%\begin{figure*}[htp]%[!h]%[htp]
	\vskip -0.06in
	\centering
	
	\includegraphics[width=0.5\textwidth]{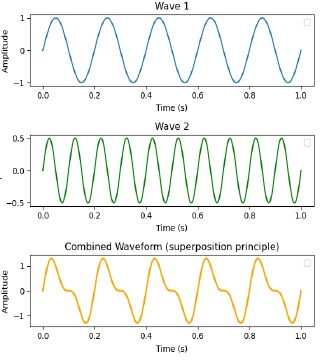}
	
	\caption{The linear combination of two waves. {\small{Note: When linearly combining two waves, the waves combine constructively, resulting in a stronger single wave. This way, the combination of a language-based metric and a vision-based metric can result in a possible stronger metric. This process is the Fourier transform.}} }
	
	\label{fig:wave}
	\vskip -0.11in	
	%\end{figure*}
\end{wrapfigure}
The second related metric, ``overall\_semsim'', is an extension of the ``sentence similarity'' calculation, designed to evaluate the contextual relevance of an object name within a text while also considering its semantic relationship with other object names in one image. Initially, the ``sentence similarity'' is computed between the object name and the caption. This value serves as the baseline for the overall similarity. We then calculate the semantic relationships between the object name and other object names present in the image using the \texttt{cosine\_similarity} method. The object names can be found in a pretrained dataset of word vectors. For example, ``baby'' has its word vector in the pretrained dataset (\url{https://fasttext.cc/docs/en/english-vectors.html}), and ``lemon'' also has its word vector in the dataset. Using the cosine method, we can obtain the cosine similarity between ``baby'' and ``lemon''. The similarity scores between the word of the target object and any words of its surrounding objects are summed to get a value called ``words\_sim'', representing the semantic relation between this object and its surrounding objects. Further, ``words\_sim'' is added to the baseline sentence similarity, resulting in the ``overall sentence similarity''. This comprehensive metric captures both the textual context and inter-object semantic relationships, making it particularly useful for tasks involving multi-object scenarios, such as image captioning or contextual object recognition. The panel \textbf{C} of Fig.~\ref{fig:semrev} illustrates how these language-based metrics are computed.

We believe that vision-based metric could co-work with language-based metric, that is, the two types of metric could be linearly incorporated to yield a stronger metric. This is like the Fourier transform.  The Fourier transform allows several different effects to be combined. For instance, the Fourier transform is a linear operation, meaning that the transform of a sum of functions is the sum of their transforms. This property allows different effects or signals to be combined and analyzed together in the frequency domain. By representing a signal as a sum of sinusoidal components, the Fourier transform enables the analysis of each component separately. Fig.\ref{fig:wave} demonstrates the inverse process of Fourier transform where two components could be transformed into a signal. This is particularly useful in signal processing, where different frequency components can be manipulated independently before being recombined \citep{sun2023optimizing}. In the similar way, we can transform one vision-based metric and language-based metric into a new metric by linearly combing them: ``sum\_vissem\_sim'' = ``overall\_semsim'' + ``obj\_image\_vissim''.  The panel \textbf{D} of Fig.~\ref{fig:semrev} illustrates how the vision-based metric is integrated with the language-based metric.

Ultimately, for comparison and reference, we also adopted the method of \citet{hayes2021looking} to compute semantic relevance using the newly updated \texttt{ConceptNet Numberbatch}\citep{speer2017conceptnet}. The computation method is the sum of the cosine similarity values among the word of the target object and the words of its surrounding objects, termed ``concepts\_sim''. The computation method is the same as our ``words\_sim'', but the difference is that we used the pretrained database of word vectors, while ``concepts\_sim'' is based on the pretrained vector database of \texttt{ConceptNet Numberbatch}. Note that \citet{hayes2021looking} used manual annotations to obtain the object names in one image. In contrast, the current study employed the \texttt{DETR} model (\texttt{facebook/detr-resnet-50}) to identify objects and obtain their names, which is another difference from \citet{hayes2021looking} . All metrics and their meanings are summarized in Table~\ref{table:sem}. Overall, \textbf{all of these computational metrics could represent both individual object semantics and its semantic relationship to other scene elements}.

%\vspace{-1.5mm}

Moreover, the visual system is hierarchically organized, with specialized anatomical areas for different processing functions. Low-level processing focuses on retinal image contrast, while high-level processing integrates diverse information sources into conscious visual representations. The tasks in the ``Human Attention in Image Captioning'' dataset involve higher-level visual processing, which relies on both top-down and bottom-up processes. This hierarchical structure allows for efficient processing of complex visual information, from basic features to more abstract concepts. Some of vision-based metrics (e.g., ''objs\_vissim'') effectively simulate the bottom-up processing hypothesis in visual perception. Bottom-up processing involves the analysis of sensory input based on inherent features such as color, shape, and contrast \cite{dijkstra2017distinct}. By incorporating these low-level visual features, our computational metrics capture the rapid, automatic processing of the bottom-up visual information. In contrast, these language-based metrics effectively simulate the top-down processing hypothesis in visual cognition. Top-down processing uses prior knowledge, expectations, and context to interpret sensory information  \cite{gilbert2013top}. The combined metric integrates both top-down and bottom-up processing approaches. By comparing the predictive power of these metrics, we can assess which types of processing are dominant in various visual processing tasks. This comparative analysis provides insights into the relative contributions of semantic knowledge and perceptual features in shaping visual perception and attention allocation. % Our findings suggest that a balanced interaction between top-down and bottom-up processes is crucial for a comprehensive understanding of human visual cognition in real-world scenarios.

\subsection{Computing saliency}

We computed visual saliency maps using spectral residual saliency. The method is designed to highlight regions of an image that are likely to attract human attention. %Regarding the \texttt{gaussian\_saliency} metric, we first generated a Gaussian saliency map by first converting the input image to grayscale and then applying a Gaussian function centered at the image's center. The Gaussian function is defined as \(\text{gaussian}(x, y) = \exp\left(-\frac{x^2 + y^2}{2\sigma^2}\right)\), where \(x\) and \(y\) are the coordinates relative to the center of the image, and \(\sigma\) is the standard deviation controlling the spread of the Gaussian. The resulting saliency map is normalized to have a maximum value of 1.
The \texttt{spectral\_residual\_saliency} requires to compute saliency based on the spectral residual method. It begins by converting the image to grayscale and computing its 2D Fourier transform, yielding the amplitude and phase spectra. The logarithm of the amplitude spectrum is smoothed using a Gaussian filter, and the spectral residual is obtained by subtracting the smoothed log amplitude from the original log amplitude. The inverse Fourier transform of the exponential of the spectral residual combined with the phase spectrum is computed to obtain the saliency map. The saliency map is then normalized to a range of [0, 1] by subtracting its minimum value and dividing by its range. These saliency maps can be used in various computer vision applications, such as object detection, image segmentation, and attention modeling.

%\subsection{Other control predict}

%Word frequency, the index of age of acquistion (AoA), 

\subsection{Statistical method}
We employed Generalized Additive Mixed Models (GAMMs) \citep{wood2017generalized} to investigate the predictive power of the metrics of our interest on fixation measures.  The present study included these control predictors (the porportion of object in an image, the saliency of the object in this image,  and the two metrics have been extensively in the related research to show that both have stronlgy influence human visual perception and processing) and random variable (e.g., ``participants'').  Including these variables is crucial for achieving optimal GAMM fitting.

GAMMs are effective in analyzing nonlinear effects and multiplicative interactions between variables, making them ideal for evaluating the predictability of semantic similarity. They are more flexible than traditional regression methods in modeling complex relationships between variables. %Eye-tracking data is simpler to analyze statistically than EEG and fMRI data, which makes it an ideal choice for our study on naturalistic discourse reading. 
Assessing model performance and comparing models can be challenging, and relying solely on correlations can be limiting. Fortunately, GAMMs are well-suited for comprehensive and precise assessments of model performance. We compared models using \texttt{AIC} (Akaike's Information Criterion) values, where a smaller value indicates a better model.

\texttt{AIC} or \texttt{BIC} (Bayesian Information Criterion ) are both measures of model fit that balance goodness of fit with model complexity. Lower values of \texttt{AIC} or \texttt{BIC} indicate better model fit. However, \texttt{AIC} is a popular criterion for comparing GAMMs, and it has some advantages over other criteria. \texttt{AIC} is designed to balance the trade-off between model fit and model complexity, penalizing models with more parameters. This makes it useful for selecting models that provide a good balance between fit and complexity. \texttt{AIC} is also relatively easy to compute and widely used in statistical modeling.

Moreover, the developer of R package on GAMM (``mgcv'') used AIC to make model comparison (\citealp{wood2016smoothing}; \citealp{wood2020inference}). AIC has also been mostly taken to understand model performance in psychological research (\citealp{wieling2018analyzing}; \citealp{baayen2020introduction} and the relevant studies) if GAMM or generalized mixed-effect models are employed.

In the studies conducted by \citet{wilcox2020predictive} and \citet{oh2023does}, the relationship between model perplexity and \texttt{$\Delta$LogLik} (log-likelihood) was utilized to analyze the perceptual competence of surprisal generated by different LMs. Their objective was to determine which LMs were capable of generating more powerful surprisal based on various corpora. In contrast, the current study aims to assess the predictive performance of our algorithms. 

A typical GAMM fitting should include control predictors. The past research show that the ``proportion of the object'' in one image play a crucial role in visual processing. Moreover, ``saliency'' could be also taken as control predictor.  The current study includes some random variables, such as ``participants'', and ```position of the object'' (i.e.,  nine categories: bottom center, bottom left,  bottom right, center,  center left, center right,  top center, top left, top right)
%We should point out some potential issues when using \texttt{$\Delta$LogLik} as a criterion. \citet{wilcox2020predictive} utilized GAMMs for their analysis but did not incorporate any \textbf{random variables} in these models. Consequently, confirming the optimality of these GAMMs becomes challenging. On the other hand, \citet{oh2023does} mentioned that they employed LMER and included random effects. However, the LMERs used in \citet{oh2023does} did not include ``word frequency'' as a control predictor. It is well-established in psycholinguistics and cognitive science that both ``word length'' and ``word frequency'' are significant variables in predicting human language processing measures (e.g., reading time, ERP data). Including these two factors as \textbf{control predictors} is commonly practiced when studying human language comprehension. Meanwhile, an optimal model should also include random variables. Failure to do so may result in suboptimal GAMMs or LMERs for studying reading time. All the metrics and the statistical methods are summarized in Table~\ref{table:sem}.  %Moreover, The excessive 
\begin{table}[ht]
	\centering
	\caption{The $\Delta$ AIC on GAMM fittings with random smooths (Note: a smaller $\Delta$ AIC indicates better performance).}
	\label{table:sem}
	\adjustbox{width=\textwidth}{%
		\begin{tabular}{lcll}
			\toprule
			\textbf{Metric (abbr.)} & \textbf{Method} & \textbf{Meaning} & \textbf{Statistical Analysis} \\
			\midrule
			\makecell{obj\_image\_vissim \\ (\textcolor{red}{vision-based})} & 
			$\cos\left(\text{em}(\text{obj}), \text{em}(\text{image})\right)$
			& \makecell{The visual similarity between \\ the target object and the whole image} & GAMM \\
			\midrule
			\makecell{objs\_vissim \\ (\textcolor{red}{vision-based})}  & 
			$\sum \cos\left(\text{obj}_t, \text{obj}_c\right)$
			& \makecell{The cumulative visual similarity among \\ the object and its surrounding objects} & GAMM \\
			\midrule
			\makecell{overall\_vissim\\ (\textcolor{red}{vision-based})}  & 
			$\begin{aligned}
				&\cos\left(\text{em}(\text{obj}), \text{em}(\text{image})\right) \\ 
				&+ \sum \cos\left(\text{obj}_t, \text{obj}_c\right)
			\end{aligned}$
			& \makecell{The sum of ``obj\_image\_vissim'' \\ and ``objs\_vissim''} & GAMM \\
			\midrule
			\makecell{sent\_semsim \\ (\textcolor{blue}{language-based})}  & 
			$\text{SentSim}(\text{obj\_name}, \text{caption})$
			& \makecell{The sentence similarity between \\ the object name and the caption} & GAMM \\
			\midrule
			\makecell{words\_semsim\\ (\textcolor{blue}{language-based})}  & 
			$\sum \text{sim}\left(\text{obj\_name}_t, \text{obj\_name}_c\right)$
			& \makecell{The sum of semantic similarity \\ among the object name and the names \\ of its surrounding objects \\ based on word2vec database} & GAMM \\
			\midrule
			\makecell{concepts\_semsim\\ (\textcolor{blue}{language-based})}  & 
			$\sum \text{sim}\left(\text{obj\_name}_t, \text{obj\_name}_c\right)$
			& \makecell{The sum of semantic similarity \\ among the object name \\ and the names of its surrounding \\ objects based on ConceptNet Numberbatch} & GAMM \\
			\midrule
			\makecell{overall\_semsim\\ (\textcolor{blue}{language-based})}  & 
			$\begin{aligned}
				&\text{SentSim}(\text{obj\_name}, \text{caption}) \\
				&+ \sum \text{sim}\left(\text{obj\_name}_t, \text{obj\_name}_c\right)
			\end{aligned}$
			& \makecell{The sum of sent\_semsim \\ and words\_semsim} & GAMM \\
			\midrule
			\makecell{sum\_semvis\_sim\\ (\textcolor{green}{combined})}  & 
			$\text{``overall\_semsim''} + \text{``obj\_image\_vissim''}$
			& \makecell{The integration of one language-based metric \\ and one vision-based metric} & GAMM \\
			\bottomrule
	\end{tabular}}
\end{table}

\section{Results}

\subsection{GAMM fittings with random effects}

We fitted 16 GAMM models to analyze the metrics of our interest (seven metrics) as the main predictors of the response variable (i.e., total duration, and fixation number). The GAMM models include \textit{object proportion} and \textit{object saliency}. \texttt{Position} and \textit{participant} are included as random variables.  The base model is like:
\\
 \texttt{
 	\hspace*{1cm} log(total duration) (or log(fixation number)) $\sim$ \\
 	\hspace*{1cm} s(proportion) \\
 	\hspace*{1cm} + s(saliency) \\
 	\hspace*{1cm} + s(participants, bs=``re'') \\
 	\hspace*{1cm} + s(position, bs = ``re'')
 }
 (here, \texttt{s} = smooth; \texttt{re} = random effect ).  This is what an optimal GAMM formula incorporating the metric of our interest looks like:
 \\
 \texttt{
 	\hspace*{1cm}log(total duration) (or log(fixation number)) $\sim$ \\
 	\hspace*{1cm} s(proportion) + \\
 	\hspace*{1cm} s(saliency) + \\
 	\hspace*{1cm} s(metric) + \\
 	\hspace*{1cm} s(participants, bs=``re'') + \\
 	\hspace*{1cm} s(position, bs = ``re'')
 }

The six metrics of interest were individually subjected to GAMM fittings. The dependent variable was log-transformed to approximate a normal distribution, thereby improving the model fit. Our analysis employs GAMMs to investigate the significance of various metrics of our interest on total duration or fixation number. We consider a variable significant when its \textit{p}-value is less than 0.05. The results are reported as follows: 1) The ``concepts\_sim'' metric is not significant in two GAMM fittings with the response variables (fixation duration and fixation number). Even when the two control predictors are removed and either of the two random variables remains, this ``concepts\_sim'' metric remains insignificant. In contrast, the ``words\_sim'' metric is significant in both GAMM fittings. As we know, these two metrics have the same computational methods but were computed based on different pretrained databases of vectors. This suggests that the sum of semantic similarity based on \texttt{ConceptNet Numberbatch} may not be as effective as the ``words\_sim'' metric based on the word2vec database. 2) We found that the two control predictors and the two random variables are significant across all GAMM fittings. 3) The other six metrics we proposed are significant in all GAMM fittings. 

In order to compare the performance of these metrics, we emloyed the $\Delta$\texttt{AIC} of different GAMM fittings to gain insight. The \texttt{AIC} of the GAMM fitting with metric subtracts form the \texttt{AIC} of the base model. When the resulting $\Delta$\texttt{AIC} is negative, it suggests that the metric in this GAMM fitting makes substantial contribution. When $\Delta$\texttt{AIC} is smaller, it indicates that the GAMM fitting with the metric contributes more, that is, the performance of this metric is better. Put it simply, a smaller $\Delta$\texttt{AIC} is indicative of better performance of this metric.  Table~\ref{table:total} summarizes the GAMM fitting results.

\begin{table}%[ht]
	\centering
	\caption{The $\Delta$ AIC on GAMM fittings with random effect.(Note: a smaller $\Delta$ indicates better performance. n = 9538)}
	\vskip 0.09in
	\label{table:total}
	\resizebox{1.08\textwidth}{!}{%
		\begin{tabular}{||c c c c c c c c||} 
			\hline
				\textbf{GAMMs} & \makecell{\textbf{obj\_}\\\textbf{image\_}\\\textbf{vissim}} &  \makecell{\textbf{objs\_}\\\textbf{vissim}} & \makecell{\textbf{overall\_}\\\textbf{vissim}} & \makecell{\textbf{sent\_}\\\textbf{semsim}} & \makecell{\textbf{words\_}\\\textbf{semsim}} & \makecell{\textbf{overall\_}\\\textbf{semsim}} & \makecell{\textbf{total\_}\\\textbf{vissem\_}\\\textbf{sim}} \\
			\hline\hline
			
			\makecell{total\\ duration}  & -317.73 & -234.1 & -265.08 &-302.27 & -455.89& -513.63 & \textcolor{blue}{-573.04} \\
			\hline\hline
			\makecell{fixation\\ number} & -304.19& -254.72& -259.29& -272.7& -449.51& -502.48 & \textcolor{blue}{-559.2} \\
			
			\hline\hline
		\end{tabular}%
	}
	\vskip -0.09in
\end{table}

These findings provide valuable insights into the roles of the metrics in human visual processing. % The consistent significance of word frequency across all channels suggests its fundamental role in word recognition and retrieval. The variable significance of word length, particularly in the Frontal region, may reflect differences in processing strategies or cognitive load associated with word size.
All these metrics are significant because $\Delta$\texttt{AIC} is negative. We found that: \textit{total\_vissem\_sim$>$overall\_semsim $>$ words\_semsim $>$ obj\_image\_vissim $>$ sent\_semsim $>$overall\_vissim $>$ objs\_vissim}. 
The superior performance of ``total\_vissem\_sim'', as indicated by AIC, underscores the importance of the metric in predicting human visual processing. It suggests that the metrics related with the language-based method shows the super predicative power but the metric incorporating the information on language and vision has the best performance. %The widespread significance of surprisal across most brain regions aligns with predictive processing theories of language comprehension, suggesting that the brain continuously generates and updates predictions during language processing.

\begin{sidewaysfigure}
	\vskip -0.06in
	\centering
	
	\includegraphics[width=1.1\textwidth]{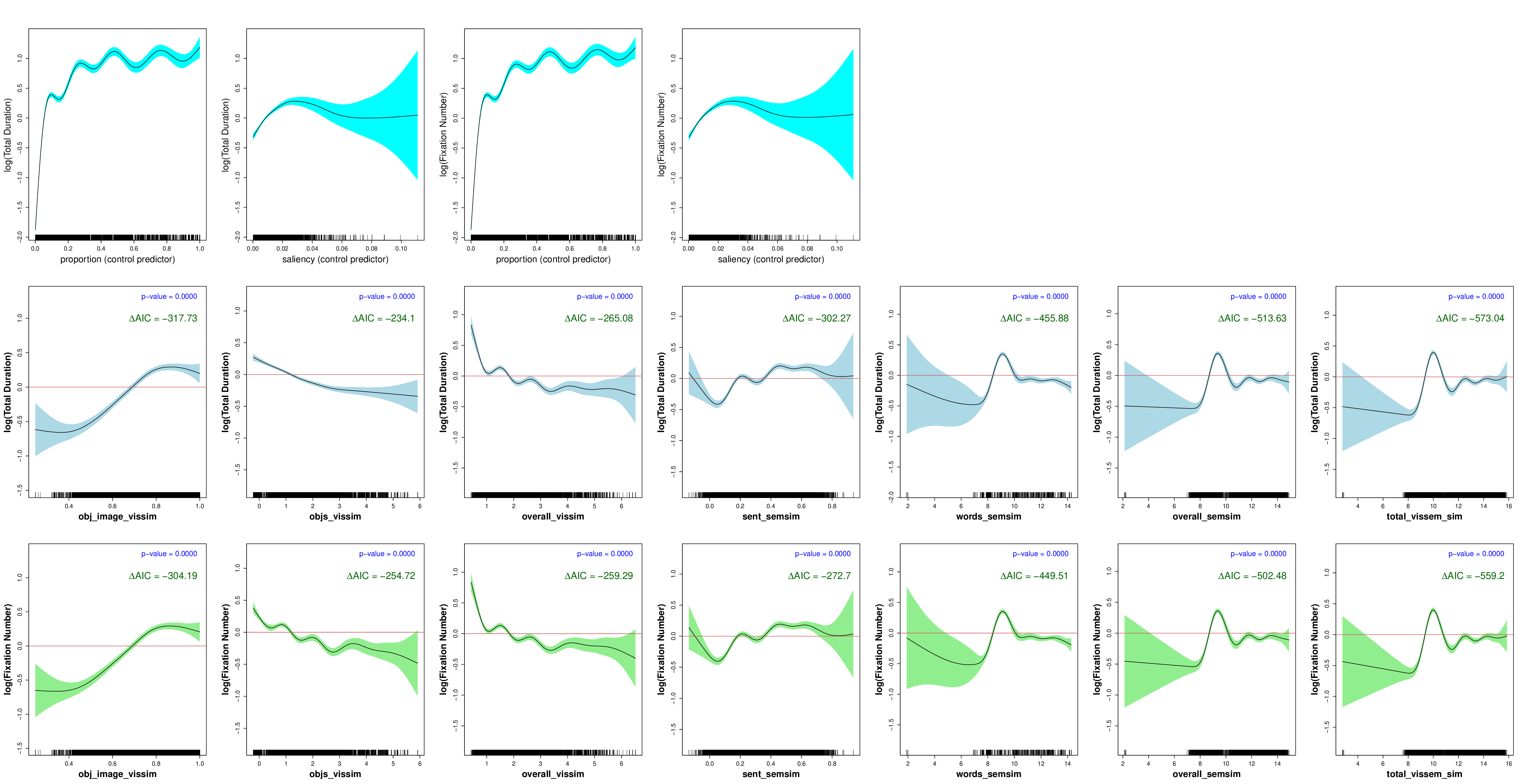}
	
	\caption{The partial effects of main metrics. Note: The first row displays the partial effects of control predictors. The second row shows the partial effects of the metrics of interest on total duration. The third row illustrates the partial effects of the metrics of interest on fixation numbers. }
	
	\label{fig:eff}
	\vskip -0.11in	
%\end{figure*}
\end{sidewaysfigure}

Fig.~\ref{fig:eff} visualizes  the partial effects of the control predictors and the metrics of our interest. When the ``proportion'' becomes greater, humans need to make greater efforts to process this object. ``Saliency'' has an opposite effect on fixations. These are consistent with the common sense and the past research. We found that the metric of the visual similarity between object and caption outperform either the metric of visual similarity among the object and its contextual objects or the metric of sum of visual similarities. As the increase of the metric of visual similarity between object and caption, its influence is positive to fixation measure. In contrast, the visual similarity among the object and its contextul objects has a negative impact on the fixation measure. The impact trend of ``objs\_vissim'' is distinct from the one of the other metrics. 

%\begin{figure*}[htp]%[!h]%[htp]

The $\Delta$\texttt{AIC} values in Table~\ref{table:total} suggest that language-based metrics outperform vision-based metrics. However, the visualization of partial effects in Fig.~\ref{fig:eff} reveals a more unique picture. The vision-based metrics (left three plots in the second and third rows) demonstrate clear, consistent trends. In contrast, the language-based metrics (right three plots in the second and third rows) exhibit more complex patterns. For instance, the ``words\_semsim'' plot shows a particularly intricate relationship. The curve initially decreases as semantic relevance increases, then sharply rises around a value of 8, followed by a rapid drop and subsequent flat fluctuation. This complex trend suggests that language-based metrics may have a multifaceted impact on fixation duration and frequency during visual processing. While language-based metrics appear statistically superior according to $\Delta$\texttt{AIC}, their complex effects on visual processing are less straightforward to interpret compared to the more transparent trends of vision-based metrics. %This complexity highlights the need for careful consideration when applying and interpreting language-based metrics in visual processing studies.

%In terms of $\Delta$\texttt{AIC}, the language-based metrics seem to perform better, as shown in Table~\ref{table:total}. However, the visualization of the partial effects of these metrics, as shown in Fig.~\ref{fig:eff}, reveals that the trends of language-based metrics are quite complex in comparison with the trends of vision-based metrics. For example, the left three plots in the second or third rows in Fig.~\ref{fig:eff}, representing the partial effects of vision-based metrics, have a clear trends for their curves. In contrast, the right three plots in the second or third rows in Fig.~\ref{fig:eff}, representing the partial effects of language-based metrics, have quite complex trends for their curves. Regarding the plot of ``words\_semsim'', its curve falls as the semantic relevance value becomes larger. When it reaches 8, the curve dramatically rises, and then quickly drop and gradually fluctuates flatly. The trend shows that the language-based metric could have a complex impact on fixation duration or number during visual processing. In comparison with the vision-based metrics, language-based metrics are a little challenging to interpret transparently for their influence on visual processing.  

\subsection{GAMM fittings with random smooths}

We then fitted another group of 14 GAMM fittings with random smooth to analyze the six metrics as predictors of two dependent variables from the same visual eye-movement dataset. Due to the insignficant cases of ``concepts\_sim'', this metric was excluded in GAMM fittings with random smooth. The random variables are the same as in Study 1. The GAMM fittings also include \textit{proportion} and \textit{saliency} as control predictors, but \textit{position} as a ``random smooth''.  The base model is the same as the one in Study 1:
\\
 \texttt{\hspace*{1cm} log(total duration) (or log(fixation number)) $\sim$} \\
\texttt{\hspace*{1cm} s(proportion) +} \\
\texttt{\hspace*{1cm} s(saliency) +} \\
\texttt{\hspace*{1cm} s(metric) +} \\
\texttt{\hspace*{1cm} s(participants, bs=``fs'') +} \\
\texttt{\hspace*{1cm} s(position, bs=``re'')} \\
\texttt{(here, \texttt{s} = smooth; \texttt{re} = random effect)}
.

An optimal GAMM fitting with random smooth is formulated as:
\\
\texttt{\hspace*{1cm} log(total duration) (or log(fixation number)) $\sim$} \\
\texttt{\hspace*{1cm} s(proportion) +} \\
\texttt{\hspace*{1cm} s(saliency) +} \\
\texttt{\hspace*{1cm} s(metric, position, bs=``fs'') +} \\
\texttt{\hspace*{1cm} s(participant, bs=``re'')} 

Here \texttt{s} = tensor product smooth, \texttt{fs} = random smooths adjust the trend of a numeric predictor in a nonlinear way, and it covers the function of random intercept and random slope; the argument \texttt{m=1} sets a heavier penalty for the smooth moving away from 0, causing shrinkage to the mean. In Study 1, ``position'' was taken as random effect (i.e., ``re''), and random effect is random slope adjusting the slope of the trend of a numeric predictor.  In contrast,  in this GAMM equation, \texttt{position} is treated  as random smooth. Random smooth leverages random slope and random intercept to fully
assess the significance of the metrics of interest at every level of the random variable. In other words, random smooth could examine random effect more comprehensively, including both random slope and random intercept. %The result is shown in Table \ref{tab:compare1}.

The dependent variable was log-transformed in order to make the data closer to normal distribution, and thus achieving better fittings. %Compared to the GAMM fittings that incorporate a random slope, the introduction of a random smooth for \texttt{position} did not change \texttt{AIC} value. These findings reveal that neither variant of word surprisal (calculated via BERT or GPT) had significant effect on any of the three examined fixation durations. This aligns with earlier results obtained from models including random slope.
We still used the threshold of \textit{p}-value $<$ 0.05 to determine the significance of variables in a GAMM fitting. The results of GAMM fittings with random smooth show that all metrics have an effect on the two types of dependent variable quite well.  The \texttt{AIC} of the GAMM fitting with metric subtracts form the \texttt{AIC} of the base model. When the resulting $\Delta$\texttt{AIC} is smaller, it indicates that the GAMM fitting with the metric contributes more, that is, the performance of this metric is better. Put it simply, a smaller $\Delta$\texttt{AIC} is indicative of better performance of this metric.  Table~\ref{table:smooth} summarizes the GAMM fitting results.%Despite this, attention-aware metrics have stronger predictability in the eye-movement data. As shown in Table \ref{tab:compare1}, attention-aware semantic relevance with weights has the smallest \texttt{$\Delta$AIC}, indicating that the metric has the strongest effect on the three eye-movement data. In addition, the tensor interaction between \textit{word length} and \textit{word frequency} has a significant effect, suggesting that both control predictors could remarkably predict eye-movements on reading. Additionally, the random effect of the \textit{participant} is strongly significant in all GAMM fittings. The results of GAMM fitting using random smooth are consistent with those using random slope. 

\begin{sidewaysfigure}
%\begin{figure*}[htp]%[!h]%[htp]
	\vskip -0.06in
	\centering
	
	\includegraphics[width=1.1\textwidth]{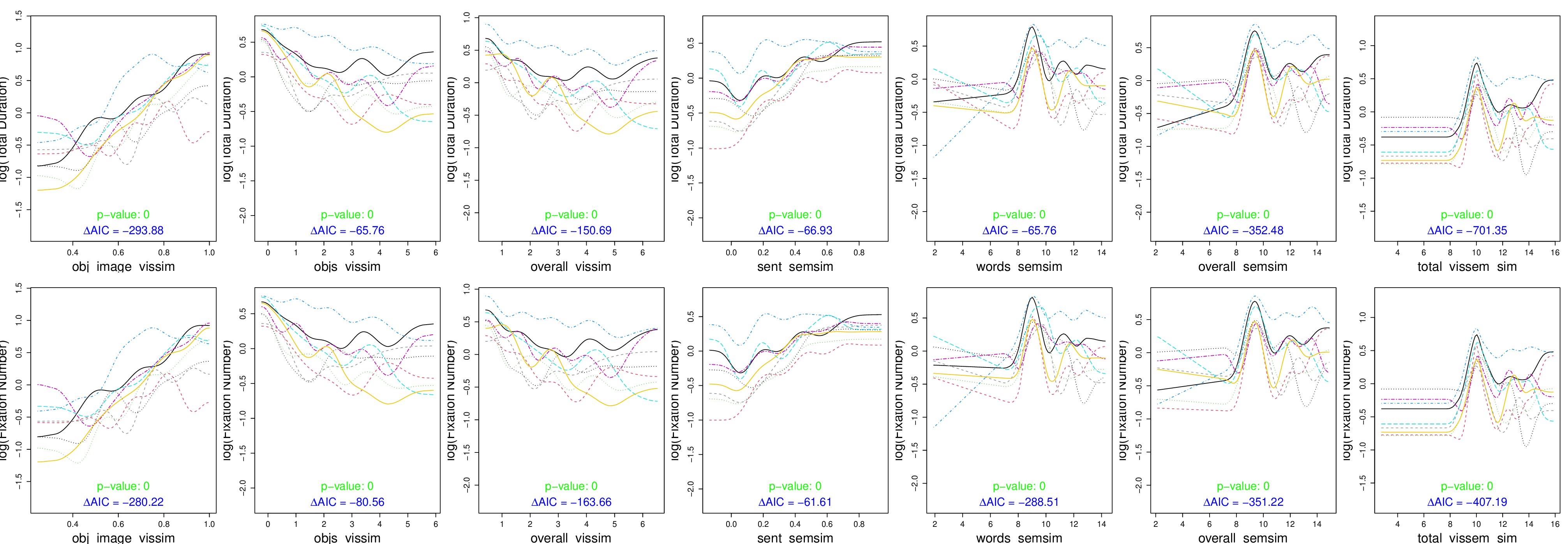}
	
	\caption{The partial effects with random smooths. The first row displays the partial effects of the metrics on total duration, while the second row shows the partial effects on fixation number. }
	
	\label{fig:smooth}
	\vskip -0.11in	
%\end{figure*}
\end{sidewaysfigure}

Fig.~\ref{fig:smooth} presents a comprehensive visualization of the effects of these metrics with random smooths.  We found that: \textit{total\_vissem\_sim $>$ overall\_semsim $>$  obj\_image\_vissim $>$ words\_semsim$>$ overall\_vissim $>$ \\ objs\_vis$>$simsent\_semsim }. 
The superior performance of ``total\_vissem\_sim'', as indicated by $\Delta$ \texttt{AIC}, underscores the importance of this combined metric in predicting human visual processing. Similarly, regarding language-based metrics, ``overall\_semsim'' has the best performance. Within the vision-based metrics, ``obj\_image\_vissim'' outperform the other two metrics, which is consistent with the results in GAMM fittings with random effect. Compared with ``obj\_image\_vissim'', ``overall\_vissim'' seems to incorporate more information on visual context, but this metric seems not predict human processing difficulty so well as ``obj\_image\_vissim'' which only consider the visually semantic relation between the object and the whole image. In contrast, ``overall\_semsim'' incorporating more language-based semantic information outperforms the other language-based metrics. Despite this, the metric combining the information on language and vision outperform other metrics.

\begin{table}[ht]
	\centering
	\caption{The $\Delta$ AIC on GAMM fittings with random smooths (Note: a smaller $\Delta$ indicates better performance. n = 9538)}
	\label{table:smooth}
	\scalebox{0.8}{
	\begin{tabular}{lccccccc}
		\toprule
		\textbf{GAMMs} & \makecell{\textbf{obj\_image\_}\\\textbf{vissim}} & \makecell{\textbf{objs\_}\\\textbf{vissim}} & \makecell{\textbf{overall\_}\\\textbf{vissim}} & \makecell{\textbf{sent\_}\\\textbf{semsim}} & \makecell{\textbf{words\_}\\\textbf{semsim}} & \makecell{\textbf{overall\_}\\\textbf{semsim}} & \makecell{\textbf{total\_vissem\_}\\\textbf{sim}} \\
		\midrule
		\makecell{total\\ duration}  & -293.88 & -65.76 & -150.69 & -66.93 & -65.76 & \textcolor{pink}{-352.48} & \textcolor{blue}{-701.35} \\
		\midrule
		\makecell{fixation\\ number}  & -280.22 & -80.56 & -163.66 & -61.61 & -288.51 & \textcolor{pink}{-351.22} & \textcolor{blue}{-407.19} \\
		\bottomrule
	\end{tabular}}
\end{table}

\section{Discussion}

\subsection{Different performance of the metrics}
Each of the six metrics we proposed demonstrates a strong ability to predict human eye movements during visual processing. These findings align with previous research on the effects of semantic similarity on visual attention. Nevertheless, the present study has new findings. The following briefly summarizes the main findings. 

%First, both vision-based and language-based metrics are remarkably significant in GAMM fittings, suggesting that both types of information are crucial for predicting human visual processing. Second, Vision-based metrics tend to have immediate negative or positive influences on fixation measures, indicating a direct relationship between visual features and attention. In contrast, language-based metrics exhibit more complex trends in predicting fixation measures. Specifically, when the language-based metric is small, it may positively influence fixation duration or number. However, as it increases, its impact could become negative. This complexity suggests that the influence of language on visual processing is not straightforward but involves intricate cognitive processes that may depend on context and the degree of semantic relevance.

First, our results indicate that both vision-based and language-based metrics take effect on human visual processing. Vision-based metrics, such as ``obj\_image\_vissim'' and ``overall\_vissim,'' showed immediate positive or negative influences on fixation measures, suggesting a direct relationship between visual features and attention. Notably, ``obj\_image\_vissim'' outperformed other vision-based metrics, indicating that the visual similarity between an object and the entire image (i.e., scene) is a strong predictor of visual processing difficulty. This aligns with previous research that emphasizes the importance of visual context in guiding attention.

Second, in contrast, language-based metrics exhibited more complex trends in their predictive power, as shown in Figs.\ref{fig:eff} and \ref{fig:smooth}. The intricate patterns exhibited by language-based metrics make their influence on visual processing more challenging to decipher compared to the clearer trends of vision-based metrics. This complexity highlights the need for careful consideration when applying and interpreting language-based metrics in visual processing studies. Moreover, these language-based metrics still vary with each greatly. For instance, the ``words\_semsim'' metric, which uses the \texttt{word2vec} pretrained database to assess semantic similarity, was significant in predicting fixation measures, while the ``concepts\_semsim'' metric, based on \texttt{ConceptNet Numberbatch}, was not. This suggests that the choice of semantic database can impact the effectiveness of language-based metrics. Additionally, the ``overall\_semsim'' metric, which integrates multiple language-based semantic information, demonstrated superior performance among language-based metrics, highlighting the nuanced influence of linguistic context on visual processing \citep{Lauer2021context}.

Third, the integration of vision and language information, as seen in the ``total\_vissem\_sim'' metric, provided the best predictive performance overall. This combined metric highlights the importance of considering both visual and linguistic information in understanding human visual processing. The superior performance of ``total\_vissem\_sim'' suggests that a holistic approach, which accounts for the interplay between visual and semantic information, is essential for accurately predicting eye movements and fixation patterns.

Our findings are consistent with past research, which highlights the role of semantic relevance in guiding visual attention \citep{jiang2023effect}. However, some inconsistencies arise, particularly in how language-based metrics influence fixation measures. These discrepancies may stem from differences in the statistical analysis setups. The following details these issues. %exploring how our results compare with existing literature and offering insights into the interplay between visual and linguistic information in shaping human perception.

%Research on semantic similarity and visual attention stressed on the significant role that semantic relationships play in guiding eye movements. 
Studies by \citet{hwang2011semantic} and \citet{hayes2021looking} show that semantic similarity among scene objects influences attention, with participants fixating more on semantically related objects during visual search tasks. Both studies employed vector-space models, such as Latent Semantic Analysis and ConceptNet, to calculate semantic relationships. \citet{wu2014guidance} advocate for integrating semantic associations into attention models, while \citet{zheng2019visual} propose methods for generating visual similarity explanations. Specifically, \citet{hayes2021looking} revealed a strong positive relationship between the semantic similarity of scene regions and viewers' focus of attention. Namely, areas containing objects semantically related to the overall scene category were more likely to capture viewers' attention. This finding is consistent with the predictability of our two metrics ( ``obj\_image\_vissim'' and ``overall\_vissim''), where the two metrics positively affect human fixation durations and numbers. These suggest that when objects share higher semantic relevance with the scene, they become focal points, enhancing visual processing efficiency.

\citet{garcia2020visual} examined how semantic relationships between objects affect visual working memory performance in healthy adults. They found that semantically related objects require more processing time when the number of objects increases, suggesting that determining whether an object was in the encoded set takes longer when objects are semantically related and the object load is higher. Our findings support this idea but diverge from the study of \citet{hayes2021looking} in some aspects. For instance, the ``objs\_vissim'' metric in Fig.~\ref{fig:eff} shows a negative effect on fixation measures, indicating that when the target object has a higher similarity with surrounding objects, less effort is needed to process the object in the image. Conversely, as the object becomes less related to contextual objects, processing becomes more difficult. A possible explanation is that humans may employ similar strategies to process objects that look quite similar, reducing cognitive load.

\citet{hayes2021looking} used object names (i.e., words) to compute semantic similarity with other object names in an image (or scene), employing \texttt{ConceptNet} to generate vectors for computation. Their method is similar to the ``words\_semsim'' proposed in our study, as both use language concepts for computation. However, the statistical analysis model (i.e., General Linear Mixed-Effects Model) \citet{hayes2021looking} employed did not include any other control predictors, which could lead the over-fitting results for the metric they used. As we know, human visual processing must be involved in a number of factors, and a number of factors (e.g., proportion, saliency etc.) work together to influence visual perception and processing. Several factors have been consistently shown to influence visual processing. Excluding established control predictors from statistical models can lead to overfitting, potentially resulting in misleading conclusions. This methodological oversight may explain why \citet{hayes2021looking} observed a strong positive relationship between the language-based semantic similarity of scene regions and viewers' attentional focus.  In our study, the trend of ``words\_semsim'' shown in Fig.~\ref{fig:eff} is more complex. When the metric is between 7 and 10, it positively relates to fixation duration, but between 10 and 14, it negatively affects fixation duration. This complexity suggests that metrics using only word vectors may lead to intricate situations. %In contrast, the ``objs\_vissim'' metric, using a similar algorithm based on vision-based vectors, shows a consistent trend in predicting eye movements, highlighting the interplay between visual and semantic processing in guiding attention.

Overall, the results of our GAMM fittings with control predictors, including both random effects and random smooths, provide a comprehensive understanding of the factors influencing visual processing. The consistent significance of control predictors such as object proportion and saliency across all models further supports their fundamental role in visual perception. By integrating insights from both vision and language metrics, our study advances the understanding of the mechanisms underlying human visual processing and offers valuable implications for the development of more effective computational models in computer vision and cognitive science.

%\subsection{Visual information working independently}
\subsection{The interplay between visual and language information}
Numerous studies demonstrate that language can affect visual perception and processing \citep{lupyan2020effects}. Our findings support these arguments, as our language-based metrics can predict human visual processing. However, the impact of language on human visual processing is complex, as shown in Figs.~\ref{fig:eff} and \ref{fig:smooth}. Moreover, research indicates that language influences both higher-level processes, such as recognition, and lower-level processes, such as discrimination and detection, often causing us to perceive in a more categorical manner. This interaction arises from the predictive and interactive nature of perception, where linguistic cues can modify how visual information is interpreted and categorized.

Despite this, \textbf{visual processing is supposed to be an independent process}, alongside language influence. The visual information-based metrics we proposed completely and effectively predicted fixation measures, indicating that visual-based semantic information plays an independent and crucial role in visual processing. However, this does not mean that language involvement is absent. We found that language-based semantic metrics could also make good predictions. However, when semantic information from language is engaged, visual processing becomes more complex, as shown the plot of ``overall\_semsim'' in Figs.~\ref{fig:eff} and \ref{fig:smooth}. This finding indicates that \textbf{the influence of language on visual processing is more complex compared to the more direct impact of visual information on visual cognition}. Moreover, the interplay between vision and language could be rather intricate and complex. For example, language can disrupt certain perceptual systems, as seen in studies where linguistic descriptions of faces affected face perception \citep{landau2010influence}. This complexity highlights the importance of considering both visual and linguistic factors in understanding human perception, as they jointly contribute to how we interpret and interact with the world around us.

The combined metric, which integrates both vision and language, consistently delivers the best performance across various cases. This finding suggests that when humans process visual information, both visual and linguistic information may be involved. The following provides further details on this point.

As shown in Tables ~\ref{table:total} and \ref{table:smooth}, the metrics derived from either vision or language alone are capable of independently predicting fixation measures, indicating that language indeed influences human visual processing. However, the metric that integrates both visual and linguistic information outperforms those based solely on vision or language in GAMM fittings. This result aligns with findings on the influence of language on visual perception \citep{lupyan2020effects}, which highlight that language effects on perception can be observed in both higher-level processes, such as recognition, and lower-level processes, such as discrimination and detection.

This newly combined metric with vision and language information is significant not only in revealing the influence of language on visual perception but also as a valuable tool for advancing our understanding of the mechanisms underlying human visual processing. By integrating visual and language information, this metric could help better understand how semantic knowledge interacts with perceptual processes. For example, this interaction is crucial because language can lead to perceiving in a more categorical manner, influencing how we interpret and respond to visual stimuli. Moreover, the integration of language cues can enhance the efficiency of visual processing by providing context and meaning, thereby facilitating quicker recognition and decision-making in complex visual environments.

\subsection{Top-down vs. bottom-up processing}
\begin{wrapfigure}{l}{0.5\textwidth}
	%\begin{figure*}[htp]%[!h]%[htp]
	\vskip -0.06in
	\centering
	
	\includegraphics[width=0.5\textwidth]{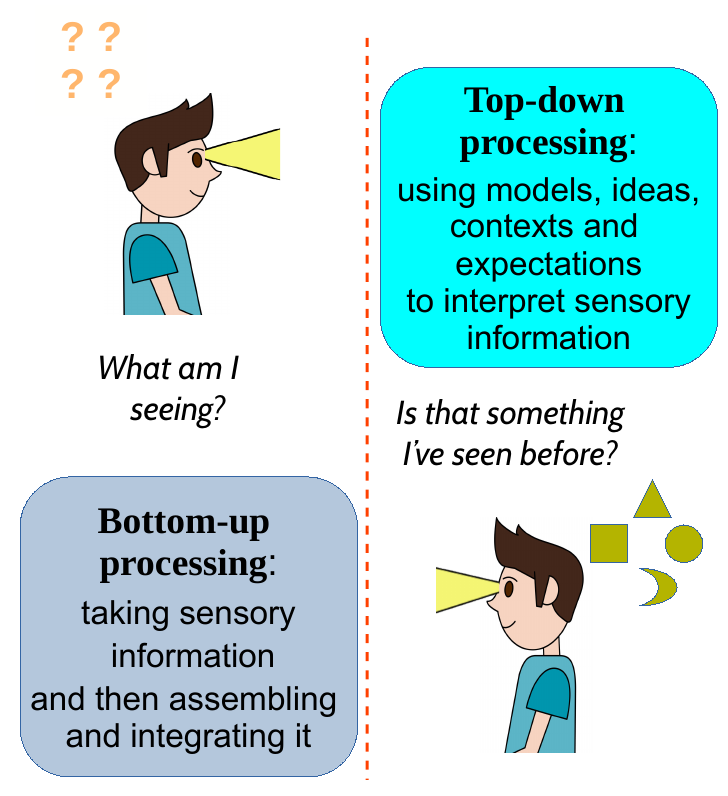}
	
	\caption{Top-down and bottom-up processing. }
	
	\label{fig:process}
	\vskip -0.11in	
	%\end{figure*}
\end{wrapfigure}

Additionally, the present study emphasizes the integration of semantic knowledge and context with vision-based features, further providing insights on the interplay between top-down and bottom-up processes in human visual processing. Top-down processing is highlighted through the role of semantic knowledge and language-based semantic relevance metrics, demonstrating how prior knowledge, expectations, and linguistic information can influence visual perception and guide attention (\citealp{rolls2008top}; \citealp{gilbert2013top}). In contrast, bottom-up processing is addressed through vision-based metrics and the consideration of visual saliency, which analyze sensory input based on inherent features like color, shape, and contrast \citep{dijkstra2017distinct}, as shown in Fig.~\ref{fig:process}. In this framework, language-based semantic relevance metrics generally reflect top-down processing, while some of vision-based metrics (e.g., ``objs\_vissim'') are more indicative of bottom-up processing. The predictability of both language-based and vision-based metrics suggests that top-down and bottom-up processing may occur simultaneously during visual processing, aligning with recent research (\citealp{connor2004visual}; \citealp{mechelli2004bottom}; \citealp{theeuwes2010top}). Despite this, the better performance of some given language-based metric (e.g., ``overall\_semsim'') indicates that top-down processing may dominate in object recongition and caption tasks and bottom-up processing may be secondary in these visual tasks. Nevertheless, the integrated metric, which combines both vision-based and language-based semantic relevance, points to a synergistic interaction between these two processes, consistent with findings in the literature (\citealp{delorme2004interaction}; \citealp{mcmains2011interactions}). This highlights the importance of both top-down and bottom-up mechanisms in achieving a comprehensive understanding of visual perception, as they jointly shape the perception and interpretation of visual scenes.

 Comparing the previous experimental methods, our research employed computational methods to estimate some given metrics which simulate top-down processing, bottom-up processing, and their combination. We evaluated these hypotheses through comparing the predictive power of these computational metrics using statistical analysis. This approach contributes to enhancing computational models in visual perception, offering a more unique understanding of how humans process and interpret visual information in real-world contexts. By integrating both top-down and bottom-up processes, our study provides valuable insights into the complex interactions between semantic knowledge and perceptual features in shaping visual cognition. The superior performance of the combined metric over individual metrics lends strong support to theories and frameworks emphasizing the intricate interplay between semantic and perceptual information in shaping visual perception. This argument shows the importance of considering both bottom-up sensory input and top-down conceptual knowledge when modeling human visual processing, potentially leading to more accurate and effective computational models of cognition.

Further, the integration of vision-based and language-based metrics into a combined metric represents a significant advancement in understanding multi-modal information processing in human cognition (\citealp{holler2019multimodal}; \citealp{benetti2023multimodal}). By merging these previously separate domains, the present study addresses a critical gap in prior studies that often treated visual and semantic similarities in isolation. This innovative approach aligns with the growing recognition in cognitive science that multi-modal processing is essential for a comprehensive understanding of human perception and cognition. 

\subsection{The predictive insight of metrics}

%Semantic relevance and word surprisal are two crucial metrics for predicting human language comprehension, each serving distinct roles with unique predictive capabilities. 
 %The following focuses on the role of semantic relevance in predicting ERP data. 
%Semantic relevance and word surprisal are two crucial metrics used to predict human language comprehension, each with distinct roles and predictive capabilities. Semantic relevance captures the broader semantic context and is particularly useful for understanding how well a word semantically fits within its local context. Semantic relevance is computed using word embeddings and can be enhanced by attention-aware methods that integrate contextual information more effectively.  More details on our approach to capture semantic information and mimic human memory in language processing is seen in \textbf{Appendix}. % Word surprisal, on the other hand, quantifies the unpredictability of a word within a given context. It is an expectation-based metric that reflects how surprising or unexpected a word is, given the preceding context. High surprisal values indicate low predictability and are associated with increased cognitive processing difficulty.

The analysis of semantic relevance metrics in predicting human visual processing has yielded several new insights, particularly through the use of GAMM with random smooths. These insights enhance our understanding of the different predictive roles of different types of semantic information influence visual attention and processing.

 First, the combined metric, ``total\_vissem\_sim,'' which integrates both vision-based and language-based information, consistently demonstrates superior predictive performance. The integration of these cues provides a more comprehensive picture of how humans interpret and respond to visual stimuli. Second, language-based metrics, particularly ``overall\_semsim,'' show complex trends in their predicting human fixation measures. This metric, which combines sentence and word-level semantic similarities, outperforms other language-based metrics. The result suggests that a holistic approach to semantic relevance, incorporating multiple layers of linguistic information, is beneficial for predicting visual processing. Third, among vision-based metrics, ``obj\_image\_vissim,'' which measures the visual similarity between an object and the entire scene, is particularly effective. This finding highlights the critical role of visual context in guiding attention and suggests that specific visual relationships are more predictive of processing difficulty than broader contextual metrics like ``overall\_vissim''. Forth, the consistent significance of control predictors such as ``object proportion'' and ``saliency'' across all models reinforces their fundamental role in visual perception. This also confirms that the two factors are control predictors. These control predictors are crucial in determining how visual information is processed and prioritized.

Overall, the metrics proposed in this study offer significant advancements in visual processing and object recognition within complex, real-world environments. By incorporating crucial contextual information, these metrics enhance object identification and categorization while improving attention allocation through the prediction of fixation measures. This contextual semantic approach, utilizing either vision or language information to compute relationships between target objects and their contexts, has proven effective in accurately predicting eye movements and fixation patterns. The success of the combined metric underscores the importance of a holistic approach that considers the synergy between visual context and semantic relevance. Our findings support the development of more interpretable and accurate computational models that integrate both visual and semantic information, addressing previous research limitations and enhancing ecological validity. These advancements contribute to a more robust understanding of visual processing in complex scenarios, with important implications for computational models in computer vision and cognitive science. The study not only extends beyond visual processing to include language comprehension and production but also opens up possibilities for multi-modal information processing in applications like augmented reality and human-computer interaction interfaces. Ultimately, this research advances our understanding of the complex interactions between visual and semantic information in shaping human perception, offering valuable insights for both theoretical research and practical applications.

In a broader sense, the methodologies and metrics we introduced have proven effective in predicting eye-movements during reading multiple languages \citep{sun2023reading}, which encompasses a facet of language comprehension. An additional inquiry in our repertoire demonstrates the efficacy of these metrics in predicting phonetic and acoustic features in spontaneous speech data. This encompasses aspects such as speech duration, intonation, pitch rate, and other acoustical dimensions \citep{sun2024differential}. Spontaneous speech is indicative of the dynamic nature of language production. Furthermore, analogous metrics introduced in another study are capable of predicting and elucidating neural activity associated with the processing of naturalistic discourse. This includes electroencephalography (EEG) and functional magnetic resonance imaging (fMRI) signals \citep{sun2024eeg}. These metrics bridge the gap between cognitive processes and their neural underpinnings. The present study substantiates that these analogous metrics and methodological approaches are not only replicable but also extensible to the realm of visual information processing in humans. In essence, these methods and metrics hold promise for deciphering and explicating the multifaceted manner in which humans process multi-modal information. %Consequently, this positions us to harness contextual semantic metrics as a unifying construct, aligning various theoretical frameworks, models, and metrics. 
This synthesis affords a more holistic and integrated perspective on the intricate cognitive and neural mechanisms underlying human multi-modal information processing.

\section{Conclusion}

The current study investigated the roles of computational metrics in elucidating the complexities of human visual processing. By leveraging deep learning techniques and a comprehensive eye movement dataset, we computed various contextual semantic relevance metrics and used them to predict fixation measures during naturalistic visual comprehension. Our findings reveal the significant predictive power of these metrics in human visual processing, highlighting the integration of visual and linguistic information as crucial to this process.  %The results underscore the effectiveness of the contextual semantic approach, which combines vision-based and language-based metrics to provide a nuanced understanding of visual processing. 
The combined metric, integrating both visual and semantic cues, demonstrated superior performance, suggesting that a holistic approach is essential for accurately predicting eye movements and fixation patterns. This integration reflects the complex interplay between predictive coding and semantic retrieval in the human brain, offering a more comprehensive understanding of how humans interpret and respond to visual stimuli. These findings contribute significantly to the field of cognitive science by enhancing our understanding of the mechanisms underlying visual perception. They also lay the groundwork for future interdisciplinary studies that could explore the implications of these metrics in educational technology and adaptive learning systems. By advancing our knowledge of how visual and linguistic information is processed, this research opens new avenues for developing more effective computational models and applications in various domains.

%\bibliographystyle{acl_natbib}
\bibliography{reference.bib}
\bibliographystyle{plain}

\appendix
%\onecolumn
%\section*{Appendix}
%\label{sec:appendix}

\end{document}